\documentclass[runningheads]{llncs}

\usepackage{eccv}

\usepackage{eccvabbrv}

\usepackage{graphicx}
\usepackage{booktabs}

\usepackage[accsupp]{axessibility}  

\usepackage[acronym]{glossaries}
\usepackage{tikz,tikz-3dplot}
\usepackage{pgfplots}
\pgfplotsset{compat=1.18}
\usepackage{subcaption}
\usepackage{float}
\usepackage{siunitx}
\usepackage{enumitem}
\usepackage{multirow}
\usepackage{makecell}
\usetikzlibrary{shapes.geometric,arrows,positioning,scopes,trees,calc,matrix,spy}

\usepackage{hyperref}

\usepackage{orcidlink}

\newacronym{vo}{VO}{visual odometry}
\newacronym{slam}{SLAM}{simultaneous localization and mapping}
\newacronym{mla}{MLA}{micro-lens array}
\newacronym{sfm}{SfM}{structure from motion}
\newacronym{rmse}{RMSE}{root mean square error}
\newacronym{sd}{SD}{standard deviation}

\begin{document}

\title{PRISM-VO: Scale-Aware Visual Odometry Using Photometric Plenoptic Bundle Adjustment} 

\titlerunning{PRISM-VO}

\author{Aymeric Fleith\inst{1,2}\orcidlink{0009-0007-5129-9195} \and
Julian Zirbel\inst{1,2}\orcidlink{0009-0006-6307-8050} \and
Daniel Cremers\inst{1}\orcidlink{0000-0002-3079-7984} \and
Niclas Zeller\inst{2}\orcidlink{0000-0001-7865-1944}}

\authorrunning{A.~Fleith et al.}

\institute{Technical University of Munich, Munich, Germany\\
\email{\{aymeric.fleith, julian.zirbel, cremers\}@tum.de}\\
\and
Karlsruhe University of Applied Sciences, Karlsruhe, Germany\\
\email{niclas.zeller@h-ka.de}}

\maketitle

\begin{abstract}
We introduce PRISM-VO, a novel pure optimization-based sparse photometric visual odometry framework for focused plenoptic cameras. The core of PRISM-VO is a novel photometric plenoptic bundle adjustment which jointly optimizes camera poses and inverse depth values of points in a sliding window. By combining geometric depth from a single plenoptic image with temporal multi-view constraints, PRISM-VO achieves accurate and drift-resilient motion estimation. Through explicit modeling of the plenoptic projection, PRISM-VO provides reliable metric-scale reconstructions, overcoming the scale ambiguity of monocular SLAM algorithms. Importantly, our approach relies solely on a single plenoptic sensor and avoids complex initialization, as depth priors are computed directly from plenoptic imaging.

Experiments show that  PRISM-VO outperforms the current state-of-the-art plenoptic visual odometry method on indoor and outdoor scenes. The proposed approach rivals other optimization- and learning-based methods while accurately and reliably recovering a metric scale of the scene.

Project page: \url{https://prism-vo.github.io/}.

\end{abstract}

\section{Introduction}
\label{sec:Introduction}

\begin{figure}[tb]
  \centering
  \includegraphics[width=0.90\linewidth]{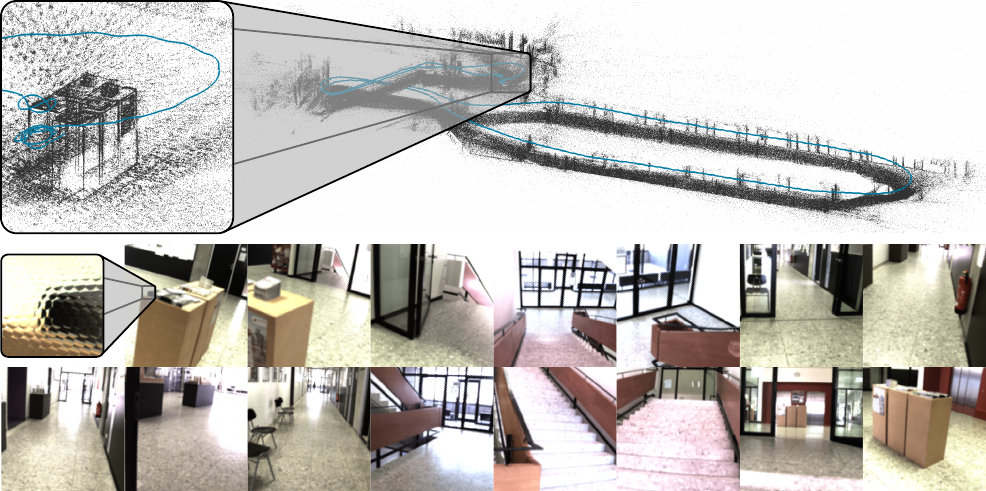}
  \caption{Metric 3D reconstruction of a 150~m long sequence (seq\_007) from the dataset~\cite{zeller2018spo} using PRISM-VO. The zoom shows the accumulated drift over the whole sequence. The images below are examples of raw plenoptic images from the sequence.}
  \label{fig:reconstruction}
\end{figure}

3D reconstruction of the environment and motion estimation are essential for real-time localization in robotics, autonomous vehicles, drones, virtual reality, and augmented reality. While lidars, radars, GPS, and inertial sensors are widely used, camera-based \gls{slam} and \gls{vo} have become widely adopted for mapping and localization. Passive cameras offer high resolution, rich visual information, versatility, and a cost-effective, lightweight, and compact sensing solution. However, monocular cameras still cannot recover absolute scale without additional information.

Alternatives such as stereo, RGB-D, and ToF cameras address the scale ambiguity of monocular vision but remain constrained by the stereo baseline, the structured light range, or the trade-off between range and accuracy. By placing a \gls{mla} between the main lens and the sensor of a camera, a plenoptic camera extends conventional imaging by simultaneously capturing spatial and angular information of the scene. This produces multiple viewpoints in the form of micro-images, also offering a very wide depth of field. This makes them well suited for true-to-scale robust \gls{slam} in compact systems.

To exploit this advantage, we propose PRISM (\textbf{P}lenoptic \textbf{R}econstruction via \textbf{I}nverse-depth \textbf{S}parse \textbf{M}apping) visual odometry, a new pure optimization-based  photometric \gls{vo} approach for focused plenoptic cameras. It combines disparity between micro-images for absolute scale estimation and larger baseline from temporal disparity to improve tracking accuracy and robustness. It enables reliable sparse metric reconstruction (see \cref{fig:reconstruction}), which is not possible with a standard monocular camera. Our main contributions are:
\begin{itemize}
    \item A sparse and photometric plenoptic bundle adjustment formulation that jointly optimizes camera poses and environment points.
    \item Tight integration of the plenoptic camera model into front-end tracking and back-end bundle adjustment of the \gls{vo} pipeline.
    \item Depth refinement using variance-weighted plenoptic depth residuals combined with temporal optimization.
\end{itemize}

We extensively evaluate the method on several datasets containing various types of scenes and different cameras. We demonstrate excellent performance compared to state-of-the-art methods by reducing the errors accumulated along the trajectory while providing a metric scale.

\section{Related Work}
\label{sec:RelatedWork}

This section reviews classical \gls{slam}, plenoptic and light-field methods, and prior \gls{sfm} and \gls{vo} approaches using plenoptic cameras.

\subsection{Classical \gls{slam} Approaches}

Early visual \gls{slam} systems relied on feature-based approaches (also called indirect methods), such as PTAM~\cite{klein2007parallel} and ORB-SLAM~\cite{mur2015orb}, which estimate camera motion by tracking sparse keypoints across frames. Later, direct methods such as DTAM~\cite{newcombe2011dtam} and LSD-SLAM~\cite{engel2014lsd} bypassed feature extraction by optimizing over image intensities. DSO~\cite{engel2018direct} further improved robustness due to photometric bundle adjustment. Methods such as DROID-SLAM~\cite{teed2021droid} rely on deep learning and achieve state-of-the-art accuracy through end-to-end optimization of camera poses and dense depth. More recently, visual–inertial \gls{slam} frameworks such as DM-VIO~\cite{von2022dm} have shown real-time performance in challenging environments.

Despite their success, these methods rely on pinhole or fisheye camera models. When applied to plenoptic cameras, their geometric and radiance assumptions no longer hold, motivating the need for adapted formulations.

\subsection{Plenoptic and Light-Field Vision}

We distinguish two main configurations of plenoptic cameras: unfocused (plenoptic camera 1.0) and focused (plenoptic camera 2.0).

Prior works studied unfocused plenoptic cameras, where the main lens is focused on the \gls{mla}. The \gls{mla} is itself focused at infinity. Calibration methods typically reconstruct sub-aperture images to detect features~\cite{dansereau2013decoding,darwish2019plenoptic,zhou2019two} or detect features directly in the micro-images~\cite{bok2017geometric,o2018calibrating,zhao2020metric}. However, they are less used today due to low spatial resolution.

Focused plenoptic cameras improve the trade-off between spatial and angular resolution by focusing micro-lenses on the main lens image plane~\cite{lumsdaine2009focused,lumsdaine2008full}. Subsequent work has established new light-field models to estimate all intrinsic, extrinsic, and scene parameters from reconstructed images~\cite{zeller2016depthcalib,zeller2017plencalib,zeller2017dpo} or directly from raw images~\cite{labussiere2020blur,labussiere2022leveraging,noury2017light}. To eliminate the need for a calibration target,~\cite{fachada2021calibration,fachada2022pattern} use sub-aperture views, and LiFCal~\cite{fleith2024lifcal} allows recalibration on any scene.

Several methods exploit light-field data for depth estimation within a single shot~\cite{wang2022occlusion,lasheras2025single}. In addition, applications such as post-capture refocusing~\cite{hahne2016refocusing}, super resolution~\cite{xiao2023cutmib} or view synthesis~\cite{fachada2025micro} highlight the versatility of light-field data.

\subsection{\gls{sfm} and \gls{vo} with Plenoptic Cameras}

Few works directly integrate plenoptic cameras into \gls{vo}/\gls{slam}. 
Early approaches proposed closed-form 6-DOF VO~\cite{dansereau2011plenoptic} or multi-scale motion estimation for low-resolution plenoptic vision~\cite{dong2013plenoptic}, while~\cite{kaveti2020light} introduced a light-field front-end for indirect \gls{slam}. However, these rely on custom setups rather than true plenoptic cameras. SPO, A semi-dense direct plenoptic odometry method, introduced in~\cite{zeller2017dpo} and improved in~\cite{zeller2018spo}, operates on micro-images and recovers metric scale from the light-field geometry. Nevertheless, this method is limited to projecting points from the penultimate image onto the last image without performing a bundle adjustment across multiple frames, which makes it less robust and more sensitive to drift.~\cite{digumarti2021unsupervised} relies on unsupervised learning for depth estimation and \gls{vo} but is tailored to sparse light-field cameras. An indirect \gls{vo} framework leveraging light-field cameras is introduced in~\cite{al5258300indirect}, which enables metric-scale translation estimation and simplifies calibration from a novel keypoint extraction. However, this approach requires an additional step of feature detection depending on textured scenes. The methods in~\cite{dury2026structure} and~\cite{fleith2024lifcal} also perform reconstruction but operate as SfM pipelines without considering temporal links between images.

Overall, light-field sensing shows strong potential for robust tracking and scale estimation. However, existing approaches target different camera setups, require training, or rely on highly textured environments.

\section{Preliminaries}
\label{sec:Preliminaries}

We first provide an overview of the plenoptic camera model used in the \gls{vo} pipeline (\cref{sec:CameraModel}). Then, we introduce the notation used in the paper (\cref{sec:Notation}).

\subsection{The Focused Plenoptic Camera}
\label{sec:CameraModel}

\begin{figure}[tb]
  \centering
  \includegraphics[width=0.90\linewidth]{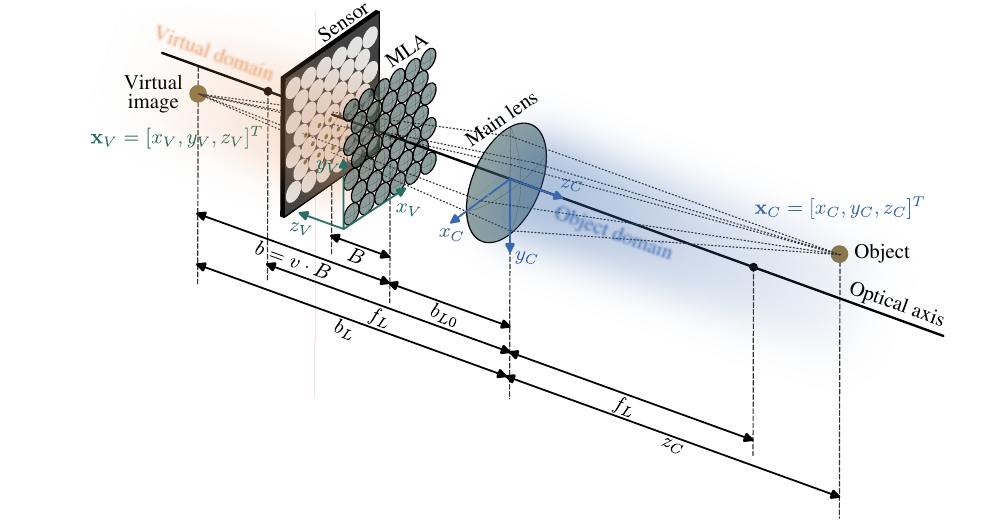}
  \caption{Representation of the model of the focused plenoptic camera in Galilean mode.}
  \label{fig:camera_model}
\end{figure}

A plenoptic camera extends the conventional imaging model by capturing not only the intensity of light rays on the sensor but also their directions within the optical system by placing an \gls{mla} between the main lens and the sensor.

Throughout the paper, we consider a focused plenoptic camera in Galilean mode and base our method on the plenoptic camera model introduced in~\cite{fleith2024lifcal}, modeling the main lens as a thin lens and the micro-lenses as pinholes. The plenoptic camera model is defined by key geometric parameters (see \cref{fig:camera_model}): $f_L$ (main lens focal length), $b_{L0}$ (distance from main lens to MLA), and $B$ (distance from MLA to sensor). The main lens forms a virtual image of the scene at a distance $b_L$ from the main lens and $b$ from the \gls{mla}. The virtual depth $v$, introduced in ~\cite{perwass2012single} as $v=\frac{b}{B}$, maps the distance $z_C$ between the real 3D point in camera metric coordinates $\mathbf{x}_C=[x_C, y_C, z_C]^T$ in the 3D virtual image space $\mathbf{x}_V=[x_V, y_V, z_V]^T$. This projection is denoted as $\Pi_{\text{pl}}(\cdot)$. The virtual depth allows a depth map to be generated without prior metric calibration of the camera. The main lens principal point is denoted as $\mathbf{c}_L=[c_x,c_y]^T$ in pixels.

\begin{figure}[t]
  \centering
  \begin{subfigure}{0.36\linewidth}
    \includegraphics[height=2.3cm]{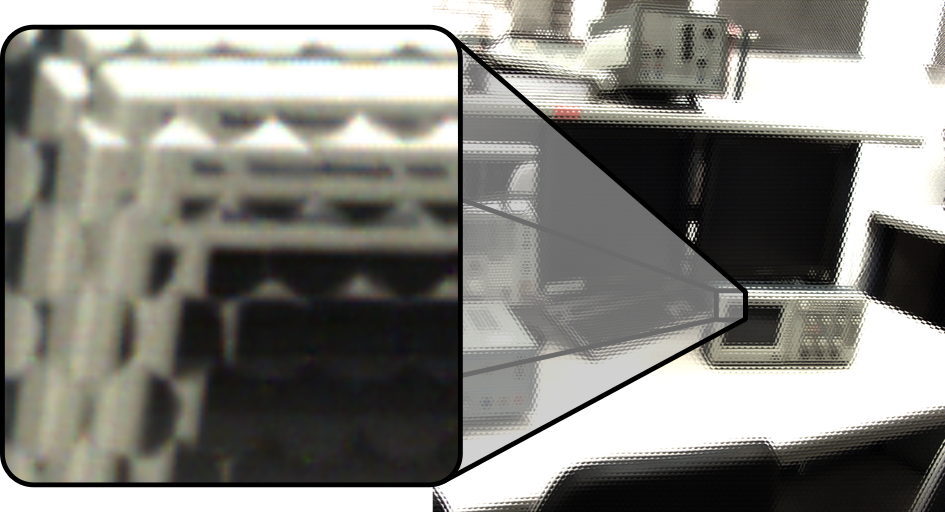}
    \caption{Raw\\image}
    \label{fig:rawImageDepthEstimation}
  \end{subfigure}
  \hfill
  \begin{subfigure}{0.20\linewidth}
    \includegraphics[height=2.3cm]{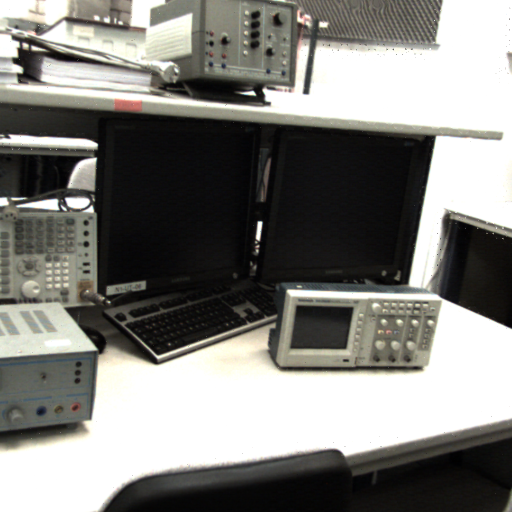}
    \caption{Totally\\focused image}
    \label{fig:totallyFocusedImageDepthEstimation}
  \end{subfigure}
  \hfill
  \begin{subfigure}{0.20\linewidth}
    \includegraphics[height=2.3cm]{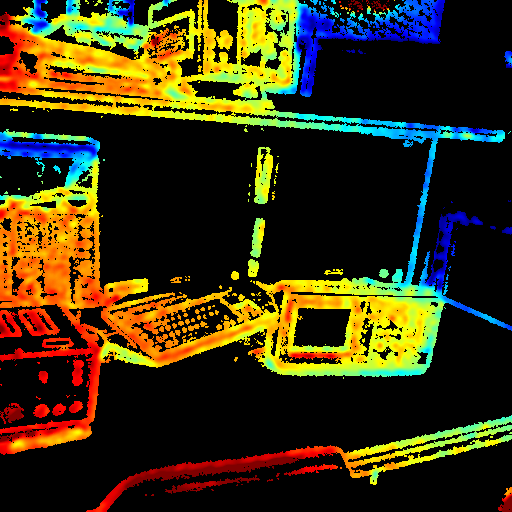}
    \caption{Virtual\\depth map}
    \label{fig:depthMapDepthEstimation}
  \end{subfigure}
  \hfill
  \begin{subfigure}{0.20\linewidth}
    \includegraphics[height=2.3cm]{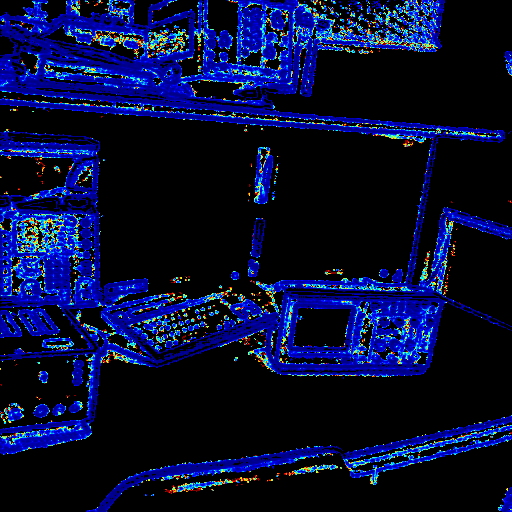}
    \caption{Virtual depth\\uncertainty map}
    \label{fig:depthVarianceMapDepthEstimation}
  \end{subfigure}
  \caption{Results of depth estimation on plenoptic images based on~\cite{zeller2015plenopticdepth} and~\cite{fleith2024lifcal}.}
  \label{fig:depthEstimation}
\end{figure}

Each micro-lens forms a small image on the sensor (see \cref{fig:rawImageDepthEstimation}). Neighboring micro-images represent the same points from slightly different perspectives. We rely on the depth-estimation pipelines of~\cite{zeller2015plenopticdepth} and~\cite{fleith2024lifcal} to compute virtual depth maps (\cref{fig:depthMapDepthEstimation}), the so-called totally focused image (\cref{fig:totallyFocusedImageDepthEstimation}) created by the main lens and the virtual depth uncertainty maps (\cref{fig:depthVarianceMapDepthEstimation}) from a single raw plenoptic image. More details about the camera model are provided in~\cite{fleith2024lifcal} and in the supplementary material.

\subsection{Notation}
\label{sec:Notation}

In the remainder of this document, vectors are denoted in bold lowercase letters ($\mathbf{x}$), matrices in bold uppercase letters ($\mathbf{J}$) and scalars in non-bold letters ($v$). Indexed quantities use subscript $i$ ($i \in \{1,\dots,n\}$) for all indexed variables, including scalars ($v_i$) and vectors ($\mathbf{x}_{C,i}$).

\section{PRISM-VO: Plenoptic Direct Visual Odometry}
\label{sec:Metod}

We first give an overview of the pipeline in \cref{sec:MethodOverview}, followed by the main components: the multi-scale image representation (\cref{sec:MultiScaleImageRepresentation}), the point selection (\cref{sec:PointSelection}), and the plenoptic bundle adjustment formulation (\cref{sec:OptimizationFormulation}).

\subsection{Method Overview}
\label{sec:MethodOverview}

\begin{figure}[tb]
  \centering
  \includegraphics[width=0.90\linewidth]{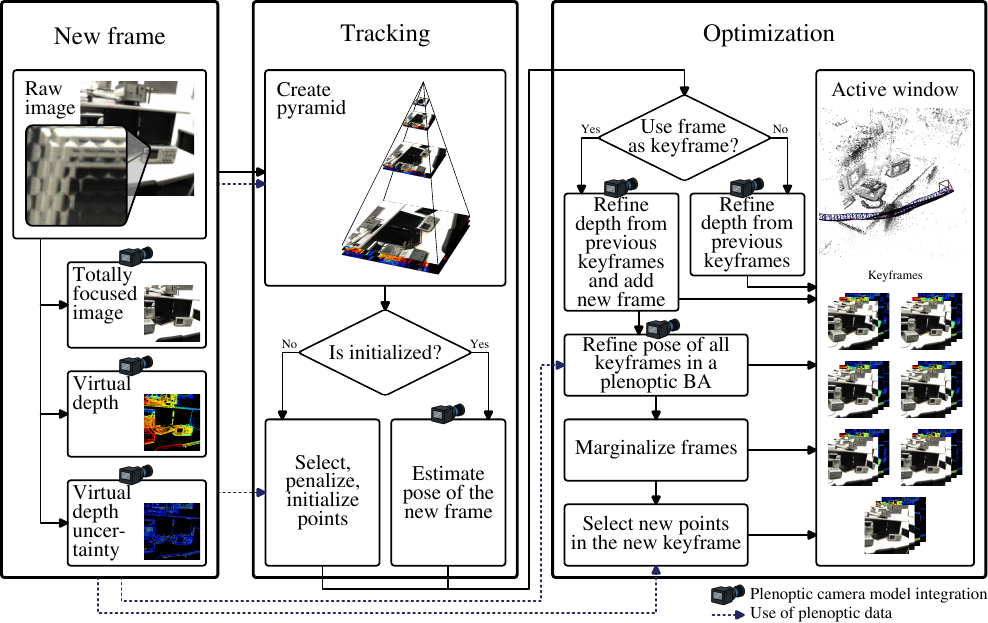}
  \caption{Overview of the PRISM-VO algorithm pipeline: image processing, tracking in the front-end, and optimization by plenoptic bundle adjustment in the back-end.}
  \label{fig:Pipeline}
\end{figure}

PRISM-VO uses a novel sparse and photometric plenoptic bundle adjustment formulation within a keyframe-based front-end/back-end architecture. It integrates the plenoptic camera model into both tracking and mapping (\cref{fig:Pipeline}). This tight coupling is required because the plenoptic camera model differs fundamentally from pinhole stereo/RGB-D models.

Each input raw image is processed into a totally focused image, virtual depth map, and virtual depth uncertainty map in a similar way as in~\cite{zeller2015plenopticdepth} and~\cite{fleith2024lifcal}. The totally focused image drives photometric tracking, while depth and uncertainty provide geometric priors for both front-end tracking and back-end optimization.

The front-end estimates camera pose via direct image alignment against reference keyframes. To increase robustness and convergence, it uses a coarse-to-fine approach. During initialization, points are selected according to their gradient and their depth information. Keyframes are added based on field of view changes, occlusions, and exposure variations in a manner inspired by~\cite{engel2018direct}.

The back-end maintains an active window of keyframes. As new keyframes are added, depth hypotheses are refined by combining multi-view photometric consistency with plenoptic depth cues. Camera poses and scene geometry are jointly optimized through a novel sparse photometric plenoptic bundle adjustment. Older keyframes are then marginalized, with their information retained in the optimization via prior constraints.

\subsection{Multi-Scale Image Representation}
\label{sec:MultiScaleImageRepresentation}

To enable robust multi-scale processing, we construct image pyramids for the totally focused image, virtual depth map, and virtual depth uncertainty map. Each pyramid level is half the resolution of the previous level in both dimensions.

\textbf{Totally Focused Image Pyramid.}
The totally focused image pyramid is built by averaging $2\times2$ pixel blocks from the previous level.

\textbf{Virtual Depth Map Pyramid.}
Since the inverse virtual depth is proportional to the micro-image disparity, one can assume it to be normally distributed. Virtual depth maps are downsampled via variance-weighted averaging, giving more weight to reliable measurements and ignoring invalid pixels. The inverse virtual depth $\bar{\rho}_{v,i}$ at the coarser level is computed with \cref{eq:rho_bar_and_sigma_rho}, where $N_x^{(i)}$ denotes the neighborhood pixels with valid inverse virtual depth $\rho_{v,i}$ and the corresponding variance $\sigma_{\rho_{v,i}}^2$.

\begin{equation}
\bar{\rho}_{v,i+1} =
\frac{\sum_{k \in N_x^{(i)}} \rho_{v,k} \cdot \left( \sigma_{\rho_{v,k}}^2 \right)^{-1}}
{\sum_{k \in N_x^{(i)}} \left( \sigma_{\rho_{v,k}}^2 \right)^{-1}}, \qquad 
\bar{\sigma}_{\rho_{v,i}}^2 =
\frac{\left|N_x^{(i)}\right|}{\sum_{k \in N_x^{(i)}} \left( \sigma_{\rho_{v,k}}^2 \right)^{-1}}
\label{eq:rho_bar_and_sigma_rho}
\end{equation}

\textbf{Virtual Depth Uncertainty Map Pyramid.}
The inverse virtual depth variance $\bar{\sigma}_{\rho_{v,i}}$ at coarser levels is computed using a harmonic mean over valid child pixels with \cref{eq:rho_bar_and_sigma_rho}, where $\left|N_x^{(i)}\right|$ is the cardinal of $N_x^{(i)}$. This approach ensures that regions with high uncertainty are appropriately reflected in coarser levels, \ie assuming a strong correlation between depths of neighboring pixels.

\subsection{Point Selection}
\label{sec:PointSelection}

\begin{figure}[t]
  \centering
  \begin{subfigure}{0.19\linewidth}
    \includegraphics[width=\linewidth]{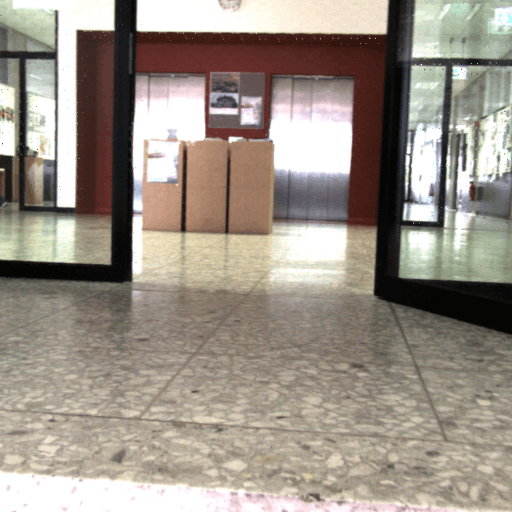}
    \caption{Totally\\focused image}
    \label{fig:totallyFocusedImage}
  \end{subfigure}
  \hfill
  \begin{subfigure}{0.19\linewidth}
    \includegraphics[width=\linewidth]{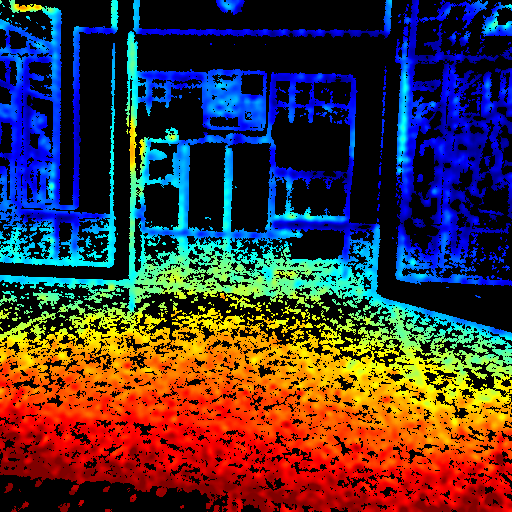}
    \caption{Virtual\\depth map}
    \label{fig:depthImage}
  \end{subfigure}
  \hfill
  \begin{subfigure}{0.19\linewidth}
    \includegraphics[width=\linewidth]{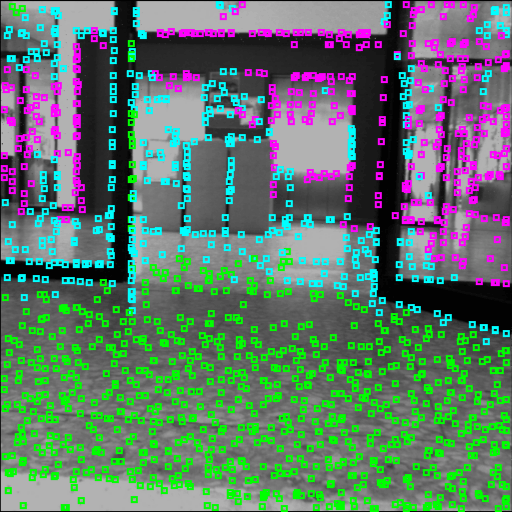}
    \caption{Point\\selection}
    \label{fig:pointSelection}
  \end{subfigure}
  \hfill
  \begin{subfigure}{0.19\linewidth}
    \includegraphics[width=\linewidth]{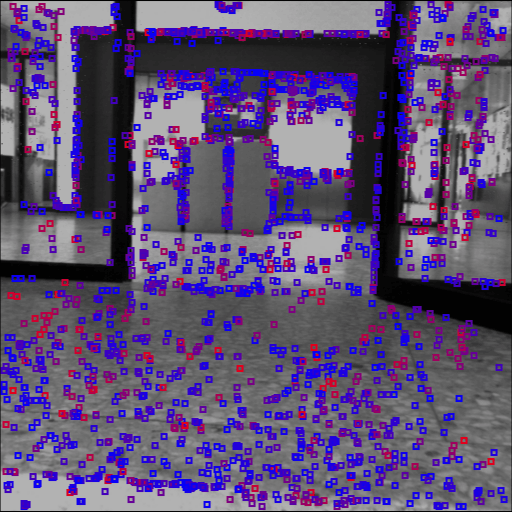}
    \caption{Photometric\\residuals}
    \label{fig:photometricResiduals}
  \end{subfigure}
  \hfill
  \begin{subfigure}{0.19\linewidth}
    \includegraphics[width=\linewidth]{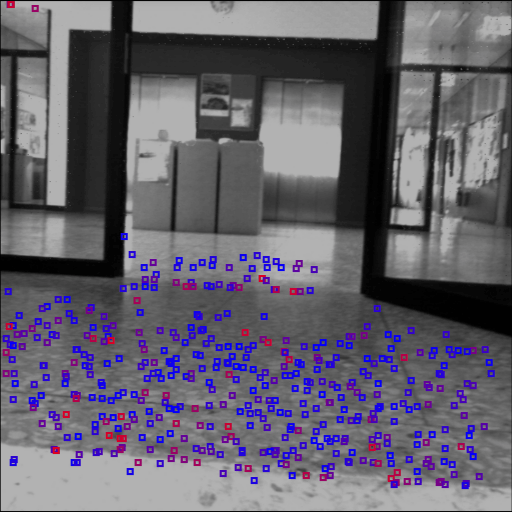}
    \caption{Inverse virtual\\depth residuals}
    \label{fig:inverseVirtualDepthResiduals}
  \end{subfigure}
  \caption{Point selection and residual definitions. (\ref{fig:totallyFocusedImage}) Totally focused image. (\ref{fig:depthImage}) Corresponding virtual depth map. (\ref{fig:pointSelection}) Selection of points when adding a new frame. \emph{Green}: valid depth used in the bundle adjustment; \emph{cyan}: selected points with depth information available but unused due to a high uncertainty; \emph{magenta}: selected points without depth. Representation of photometric (\ref{fig:photometricResiduals}) and depth (\ref{fig:inverseVirtualDepthResiduals}) residuals, where \emph{red} indicates high energy and \emph{blue} low energy.}
  \label{fig:pointSelectionAndResiduals}
\end{figure}

\textbf{Adaptive Depth Cutoff.}
We compute an adaptive inverse depth cutoff based on the mean inverse virtual depth $\rho_{v,\text{av}}$ across valid points to favor points near the camera with lower variance. Using a normalized weighting factor $w_\text{cut} \in [0,1]$, we define a percentile $p_a$ with \cref{eq:p_a} between $p_{\min}$ and $p_{\max}$ that reduces the influence of distant outliers for reliable depth selection (see \cref{fig:pointSelectionAndResiduals}). The values $\rho_{v,\text{n}}$ (near) and $\rho_{v,\text{f}}$ (far) represent reference inverse virtual depth bounds.

\begin{align}
p_{\text{a}} = p_{\min} + w_\text{cut} \cdot (p_{\max} - p_{\min}),
\qquad
\text{with}
\qquad
w_\text{cut} = \frac{\rho_{v,\text{av}} - \rho_{v,\text{f}}}{\rho_{v,\text{n}} - \rho_{v,\text{f}}}
\label{eq:p_a}
\end{align}

\textbf{Point Penalization.}
The gradient-based score $s_{\text{grad}}(\mathbf{x}_{V})$ of a point $\mathbf{x}_{V}$ is defined from the totally focused image gradient $\nabla I(\mathbf{x}_{V})$ and a randomly sampled unit direction $\mathbf{d}$ to encourage orientation diversity. The score of a pixel $\mathbf{x}_{V}$ containing valid depth is weighted more heavily by a constant factor $w_d(\mathbf{x}_{V})$ than those without depth information. The final selection score used for ranking candidate pixels is therefore computed with \cref{eq:selection_score}. Points without depth data are also selected (but with a lower weight $w_d(\mathbf{x}_{V})$) to ensure good distribution of points across the image. \cref{fig:pointSelectionAndResiduals} illustrates the resulting point selection. This approach favors reliable nearby points while ensuring coverage in regions with noisy, missing, or distant depth.

\begin{equation}
s(\mathbf{x}_{V}) = \left| \nabla I(\mathbf{x}_{V})^\top \mathbf{d} \right| \, w_d(\mathbf{x}_{V})
\label{eq:selection_score}
\end{equation}

\subsection{Optimization Formulation}
\label{sec:OptimizationFormulation}

The state vector, including camera poses, inverse depths, and brightness parameters (to account for exposure changes), is estimated by minimizing a joint nonlinear least-squares problem. A point $\mathbf{x}_{V,i}$ in the host frame $i$ is projected with \cref{eq:projection} to a point $\mathbf{x}_{V,j}$ in a target frame $j$ using the plenoptic camera model introduced in \cref{sec:CameraModel}. $\mathbf{R},\mathbf{t} \in SE(3)$ denote the relative pose and $\Pi_{\text{pl}}(\cdot)$ denotes the plenoptic projection (see supplementary material for details).

\begin{align}
  \mathbf{x}_{V,j} = \Pi_{\text{pl}}\left(\mathbf{R} \cdot \Pi_{\text{pl}}^{-1}\left(\mathbf{x}_{V,i} \right) + \mathbf{t}\right)
  \label{eq:projection}
\end{align}

The photometric residual $r^{\text{photo}}$ for a point $\mathbf{x}_{V,i}$ in a reference frame $I_i$ observed in a target frame $I_j$ is defined in \cref{eq:r_photo}. The affine brightness parameters $a_i$, $a_j$, $b_i$ and $b_j$ are inspired by~\cite{engel2018direct}.

\begin{equation}
  r^{\text{photo}} =
  \left(
  I_j \left[ \Pi_{\text{pl}}(\mathbf{x}_{V,i}) \right] - b_j
  \right)
  -
  \frac{e^{a_j}}{e^{a_i}}
  \left(
  I_i[\mathbf{x}_{V,i}] - b_i
  \right)
  \label{eq:r_photo}
\end{equation}

The depth residual is formulated in the inverse virtual depth space as defined in \cref{eq:r_depth}. Here, $\rho_v$ is the inverse virtual depth resulting from the projection and $\rho_v^{\text{meas}}$ the corresponding measured value from the depth map.

\begin{equation}
r^{\text{depth}} =
\rho_v - \rho_v^{\text{meas}}
\label{eq:r_depth}
\end{equation}

The total energy is given by

\begin{equation}
E =
\sum_k w_k^{\text{photo}} \, \| r_k^{\text{photo}} \|_\gamma
+
\eta
\sum_l w_l^{\text{depth}} \, \| r_l^{\text{depth}} \|_\gamma,
\label{eq:energy_residuals}
\end{equation}

where $w_k^{\text{photo}}$ and $w_l^{\text{depth}}$ are per-residual weights. The quadratic error of the inverse virtual depth can be minimized as it is normally distributed. The inverse virtual depth residual is weighted by the inverse virtual depth variance, resulting in $w_l^{\text{depth}} = \left( \sigma_{\rho_{v,l}}^2 \right)^{-1}$, so more reliable depth measurements have greater influence. The symbol $\|\cdot\|_\gamma$ denotes the Huber norm, and $\eta$ balances the photometric and depth residual terms.

\textbf{Adaptive Cross-Modal Weighting.}
To balance photometric and depth residuals in the joint optimization, we introduce a scalar weight $\eta$. It is adapted online by matching the average Gauss–Newton curvature, determined by $\mathbf{J}^T\mathbf{J}$, of both residual types with \cref{eq:eta_weight}.

\begin{equation}
\eta
= c \cdot
\frac{\sum_i \| J_i^{\text{photo}} \|_2^2}
     {\sum_j \| J_j^{\text{depth}} \|_2^2}
 \quad
\text{with}
\quad
J_i^{\text{photo}} = \frac{\partial r_i^{\text{photo}}}{\partial \rho_{C,i}}
 \quad
\text{and}
\quad
J_i^{\text{depth}} = \frac{\partial r_j^{\text{depth}}}{\partial \rho_{C,j}}
\label{eq:eta_weight}
\end{equation}

Here, $c$ is a constant factor and $J_i^{\text{photo}}$ and $J_j^{\text{depth}}$ are respectively the Jacobian contributions of the photometric and the inverse virtual depth residuals related to the inverse point depths.

\begin{figure}[tb]
  \centering
  \begin{subfigure}{0.82\linewidth}
\pgfplotstableread[col sep=comma]{figures/plot_weighting_factor/depthWeightFactor.csv}\weightingfactor
\colorlet{color_blue}{blue}

\begin{tikzpicture}

\begin{axis}[%
    width=0.8\linewidth,
    height=0.33\linewidth,
    scale only axis,
    xmin=0,
    xmax=1782,
    xlabel={\footnotesize Frame ID},
    ymin=1250,
    ymax=2800,
    ylabel={\shortstack{\footnotesize Inverse virtual\\residual weight factor}},
]

\addplot[color=color_blue,solid,thick]
    table[x index=0,y index=1] {\weightingfactor};
    \label{pgfplots:lifsvo}

\addplot[
    only marks,
    mark=*,
    mark size=2pt,
    color=black
] coordinates {(678,1469.7)};
\addplot[
    only marks,
    mark=*,
    mark size=2pt,
    color=black
] coordinates {(1262,2626.83)};

\node at (axis cs:683,1469.7) [anchor=west] {\scriptsize (1)};
\node at (axis cs:1267,2626.83) [anchor=west] {\scriptsize (2)};

\end{axis}
\end{tikzpicture}
\caption{Adaptive weight factor $\eta$ for the depth residual as a function of frame IDs.}
\label{fig:plot_weighting_factor}
  \end{subfigure}
  \hfill
\begin{subfigure}{0.15\linewidth}
\centering

\begin{tikzpicture}
\node[inner sep=0] (img)
    {\includegraphics[width=\linewidth]{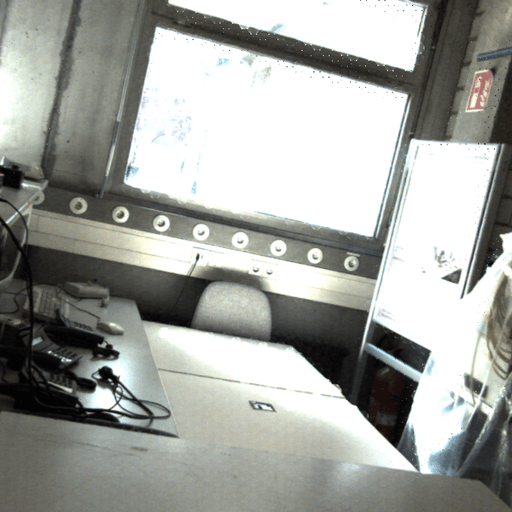}};
\node[anchor=north west, 
      fill=white,
      draw=white,
      thick,
      font=\scriptsize] 
      at (img.north west) {(1)};
\end{tikzpicture}

\vspace{1mm}

\begin{tikzpicture}
\node[inner sep=0] (img)
    {\includegraphics[width=\linewidth]{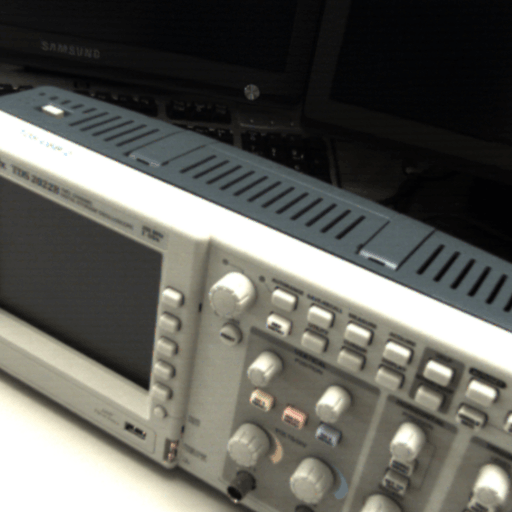}};
\node[anchor=north west, 
      fill=white,
      draw=white,
      thick,
      font=\scriptsize] 
      at (img.north west) {(2)};
\end{tikzpicture}

\caption{Totally focused images.}
\label{fig:weight_frames}
\end{subfigure}
  \caption{Evolution of the inverse virtual depth residual weight factor $\eta$ on seq\_004 from the dataset~\cite{zeller2018synchronized}. \cref{fig:plot_weighting_factor}: Local maxima occur with nearby geometry, increasing depth influence, while local minima correspond to distant scenes dominated by photometric consistency. \cref{fig:weight_frames}: Corresponding representative frames are shown.}
  \label{fig:evolution_depth_residual_weight}
\end{figure}

For temporal stability, the weight $\eta$ is updated with

\begin{equation}
\eta_{i+1}
=
\exp\!\left(
(1-\alpha)\log \eta_i
+
\alpha \log \eta
\right),
\label{eq:eta_kp1}
\end{equation}

using a fixed sensitivity parameter $\alpha$ and exponential smoothing in the logarithmic domain to decrease the weight for older values. This results in a scene-adaptive balance between photometric and depth constraints without manual tuning as shown in \cref{fig:evolution_depth_residual_weight}. Since inverse depth is more sensitive to nearby points due to stronger parallax, scenes dominated by close objects increase the scaling factor, giving depth residuals greater influence (see \cref{fig:evolution_depth_residual_weight}).

\textbf{Levenberg–Marquardt Optimization.}
The nonlinear problem is solved with a Levenberg–Marquardt optimization in a sliding window,  jointly optimizing camera poses $\boldsymbol{\xi}$ in the tangent space, affine brightness parameters $a$, $b$, and inverse depths $\rho_{C}$. At each iteration, all residuals are linearized around the current state $\mathbf{s}$ as

\begin{equation}
r(\mathbf{s} + \delta \mathbf{s})
\approx
r(\mathbf{s}) + \mathbf{J} \, \delta \mathbf{s},
\end{equation}

where $\mathbf{J}$ stacks the photometric and depth Jacobians (see the supplementary material for the complete derivation). The inverse virtual depth residual $r^{\text{depth}}$ depends only on inverse depth and does not depend on $a$, $b$ or the pose $\boldsymbol{\xi}$. The overall structure of the Hessian matrix is therefore preserved, making the integration very efficient to implement. Stacking all residuals yields the \cref{eq:gauss_newton}, where $\mathbf{H}$ is the approximate Gauss-Newton Hessian, $\mathbf{b}$ the gradient vector and $\mathbf{W}$ the weight matrix. The Schur complement is used to efficiently calculate the increment of the state vector by taking advantage of the sparsity of $\mathbf{H}$ and to marginalize keyframes that leave the optimization window, similar to~\cite{engel2018direct}.

\begin{equation}
(\mathbf{H} + \mu \mathbf{I}) \, \delta \mathbf{s}
=
- \mathbf{b},
\qquad
\text{with}
\qquad
\mathbf{H} = \mathbf{J}^\top \mathbf{W} \mathbf{J},
\qquad
\mathbf{b} = \mathbf{J}^\top \mathbf{W} \mathbf{r}
\label{eq:gauss_newton}
\end{equation}

\section{Evaluation}
\label{sec:Evaluation}

We first qualitatively demonstrate the advantages of the plenoptic camera. Then, we evaluate PRISM-VO against plenoptic-specific methods and state-of-the-art monocular \gls{vo}/\gls{slam} pipelines and  perform an ablation study of its key components. Our experiments rely on two plenoptic datasets with synchronized multi-sensor data and ground truth.
\begin{itemize}
    \item A Synchronized Stereo and Plenoptic Visual Odometry Dataset~\cite{zeller2018synchronized}. It consists of 11 indoor and outdoor sequences acquired with a Raytrix R5 plenoptic camera and a synchronized stereo camera pair, featuring large loops and depths up to several hundred meters for drift and scale evaluation.
    \item LiFMCR Dataset~\cite{FleithZirbel2025LiFMCR}. It contains 7 scenes recorded with two high-resolution Raytrix R32 plenoptic cameras, providing a precise 6-DoF ground truth from a Vicon system for precise pose accuracy assessment.
\end{itemize}

\begin{figure}[t]
  \centering
  \begin{subfigure}{0.48\linewidth}
    \includegraphics[width=\linewidth]{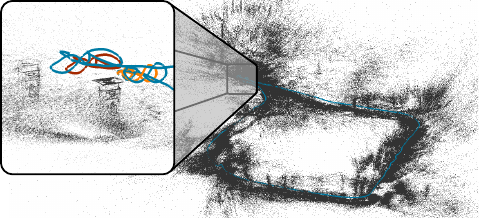}
  \end{subfigure}
  \hfill
  \begin{subfigure}{0.48\linewidth}
    \includegraphics[width=\linewidth]{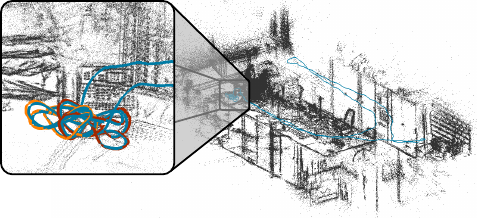}
  \end{subfigure}
  \caption{Point clouds and trajectories estimated by PRISM-VO on sequences from the dataset~\cite{zeller2018synchronized}. On the \emph{left} is the sequence seq\_002 (200~m long / outdoor) and on the \emph{right} is the seq\_009 (30~m long / indoor). The zoomed views show the accumulated drift and the ground-truth trajectory (front part in \emph{orange}, back part in \emph{red}).}
  \label{fig:point_clouds_scale_awareness}
\end{figure}

\begin{figure}[t]
  \centering
  \begin{subfigure}{0.31\linewidth}
    \includegraphics[width=\linewidth]{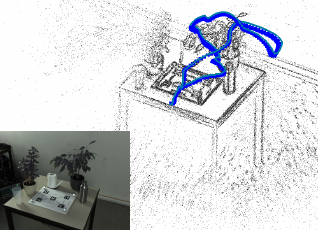}
  \end{subfigure}
  \hfill
  \begin{subfigure}{0.31\linewidth}
    \includegraphics[width=\linewidth]{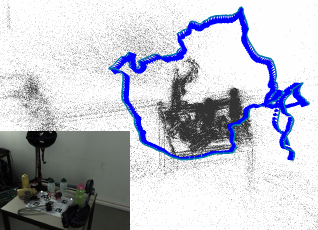}
  \end{subfigure}
  \hfill
  \begin{subfigure}{0.31\linewidth}
    \includegraphics[width=\linewidth]{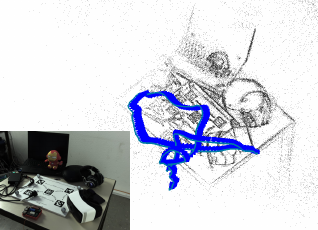}
  \end{subfigure}
  \caption{Point clouds and trajectories estimated by PRISM-VO on sequences from the LiFMCR dataset~\cite{FleithZirbel2025LiFMCR} with the associated totally focused image (scenes from left to right: 01\_Plants, 02\_Bike, 04\_Electronics). The camera poses are shown in \emph{blue}.}
  \label{fig:point_clouds_lifmcr}
\end{figure}

These datasets provide complementary scenarios for assessing long-term drift, scale consistency, and pose accuracy. Qualitative 3D reconstructions are shown in \cref{fig:point_clouds_scale_awareness} and \cref{fig:point_clouds_lifmcr} for both datasets respectively.

\subsection{Qualitative Results Against Other Sensors}

To demonstrate the benefits of the plenoptic technology, we recorded a sequence using a synchronized plenoptic-monocular rig consisting of an R32 Raytrix camera and a Basler acA1920-40gc camera. As shown in \cref{fig:trajectoryChallenging}, DSO and ORB-SLAM3 fail in a challenging fan scene, despite a wide field of view, due to parallax at intersecting grilles, transparent fan blades, and reflective surfaces. DPVO remains stable but cannot recover metric scale, unlike PRISM-VO.

We also captured the same scene from the same viewpoint using the R32 Raytrix camera and an RGB-D camera (Intel RealSense D455). While the RGB-D camera produces unstable depth, the plenoptic system succeeds (see \cref{fig:DepthMap_Comparison}).

\begin{minipage}[b]{.60\textwidth}
    \centering

    \begin{figure}[H]
        \centering
        \includegraphics[width=\linewidth]{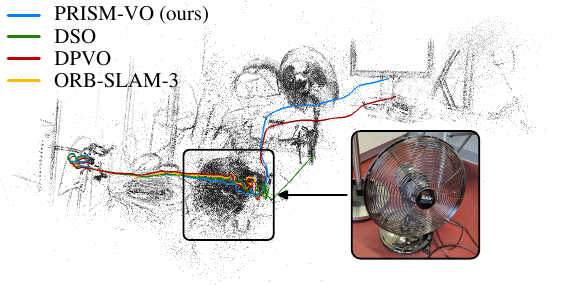}
        \caption{Trajectories from a challenging sequence where PRISM-VO succeeds while other monocular methods fail.}
        \label{fig:trajectoryChallenging}
    \end{figure}
\end{minipage}\hfill
\begin{minipage}[b]{.30\textwidth}
    \begin{figure}[H]
        \centering

        \begin{subfigure}[t]{0.47\linewidth}
            \centering
            \includegraphics[width=\linewidth]{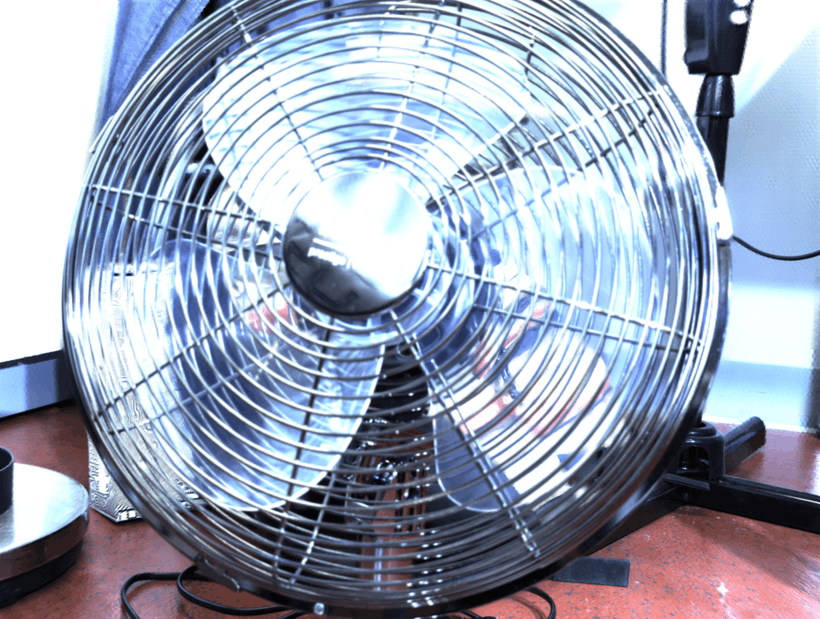}
            \caption{Focus Pl}
            \label{fig:focusPlenoptic}
        \end{subfigure}
        \hfill
        \begin{subfigure}[t]{0.47\linewidth}
            \centering
            \includegraphics[width=\linewidth]{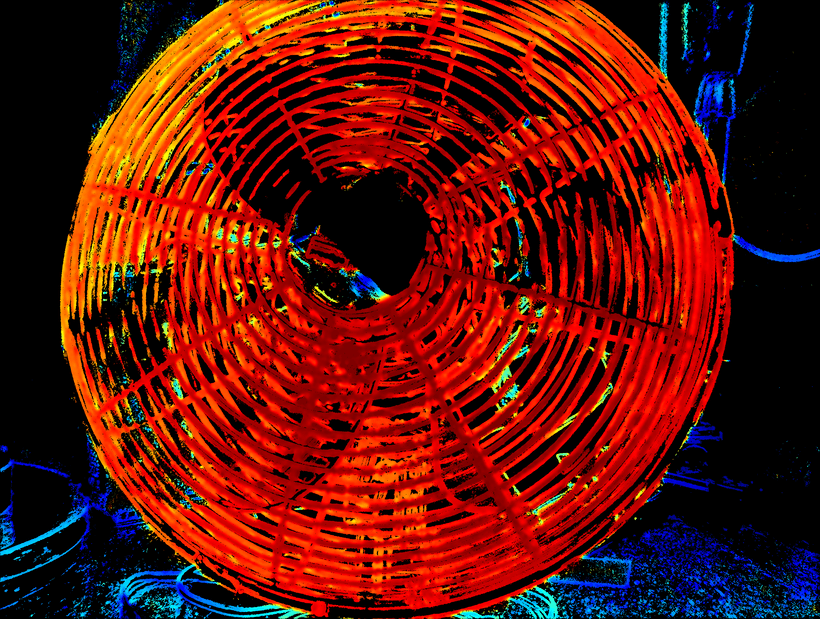}
            \caption{Depth Pl}
            \label{fig:depthPlenoptic}
        \end{subfigure}

        \vspace{0.5em}

        \begin{subfigure}[t]{0.47\linewidth}
            \centering
            \includegraphics[width=\linewidth]{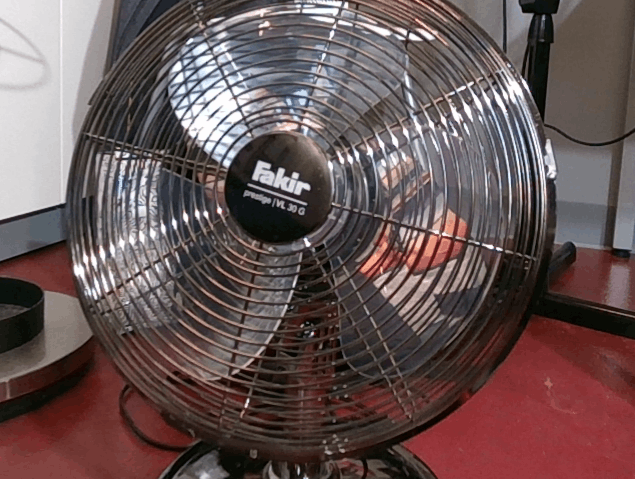}
            \caption{RGB RS}
            \label{fig:RGBRealsense}
        \end{subfigure}
        \hfill
        \begin{subfigure}[t]{0.47\linewidth}
            \centering
            \includegraphics[width=\linewidth]{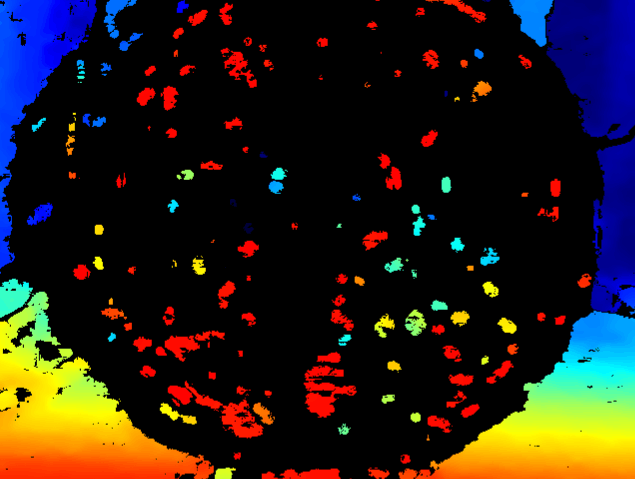}
            \caption{Depth RS}
            \label{fig:depthRealsense}
        \end{subfigure}

        \caption{Depth maps from plenoptic (\emph{Pl}) and RealSense D455 (\emph{RS}).}
        \label{fig:DepthMap_Comparison}
    \end{figure}
\end{minipage}

\subsection{Quantitative Drift Study}

To evaluate the accumulated drift over the complete \gls{vo} pipeline, we rely on the dataset from~\cite{zeller2018synchronized}. Each sequence starts and ends at the same location, forming a large loop, allowing a quantitative assessment of accumulated drift compared to the ground truth obtained by loop closure. Following~\cite{zeller2018synchronized}, we estimate two similarity transformations, $\mathbf{T}_{\mathrm{start}}^{\mathrm{gt}}$ (Eq.~\ref{eq:Tstart}) and $\mathbf{T}_{\mathrm{end}}^{\mathrm{gt}}$ (Eq.~\ref{eq:Tend}), aligning estimated 3D points $\mathbf{x}_{C,i} \in \mathbb{R}^3$ and ground-truth points $\mathbf{x}_{C,i}^{\mathrm{gt}} \in \mathbb{R}^3$ at the beginning and end of each sequence respectively. The start segment is defined by the index set $S$, and the end segment by the index set $E$.

\begin{equation}
\mathbf{T}_{\mathrm{start}}^{\mathrm{gt}} :=
\arg \min_{\mathbf{T} \in \mathrm{Sim}(3)}
\sum_{i \in S}
\left\| \mathbf{T} \mathbf{x}_{C,i} - \mathbf{x}_{C,i}^{\mathrm{gt}} \right\|_2^2
\label{eq:Tstart}
\end{equation}

\begin{equation}
\mathbf{T}_{\mathrm{end}}^{\mathrm{gt}} :=
\arg \min_{\mathbf{T} \in \mathrm{Sim}(3)}
\sum_{i \in E}
\left\| \mathbf{T} \mathbf{x}_{C,i} - \mathbf{x}_{C,i}^{\mathrm{gt}} \right\|_2^2
\label{eq:Tend}
\end{equation}

From $\mathbf{T}_{\mathrm{start}}^{\mathrm{gt}}$ and $\mathbf{T}_{\mathrm{end}}^{\mathrm{gt}}$, we compute the accumulated drift transformation $\mathbf{T}_{\mathrm{drift}} \in \mathrm{Sim}(3)$ over the entire trajectory with \cref{eq:Tdrift}.

\begin{equation}
\mathbf{T}_{\mathrm{drift}} :=
\begin{bmatrix}
e_s \mathbf{R} & \mathbf{t} \\
0 & 1
\end{bmatrix}
=
\mathbf{T}_{\mathrm{end}}^{\mathrm{gt}}
\left( \mathbf{T}_{\mathrm{start}}^{\mathrm{gt}} \right)^{-1}
=
\begin{bmatrix}
s_e \mathbf{R}_e & \mathbf{t}_e \\
0 & 1
\end{bmatrix}
\begin{bmatrix}
s_s \mathbf{R}_s & \mathbf{t}_s \\
0 & 1
\end{bmatrix}^{-1}
\label{eq:Tdrift}
\end{equation}

For evaluation, we define the following metrics: absolute scale error $d_s = \sqrt{s_e \cdot s_s}$, scale drift $e_s = \frac{s_e}{s_s}$, rotation error $e_r$ defined as the rotation angle around the Euler axis associated with the rotation matrix $\mathbf{R} \in \mathrm{SO}(3)$ and a combined alignment error $e_{\mathrm{align}}$ captures overall trajectory inconsistency (\cref{eq:ealign}). To facilitate interpretation, we define $d_s' := \max\{d_s,\; d^{-1}_s\}$ and $e_s' := \max\{e_s,\; e^{-1}_s\}$.

\begin{equation}
e_{\mathrm{align}} :=
\sqrt{
\frac{1}{N}
\sum_{i=1}^{N}
\left\|
\mathbf{T}_{\mathrm{start}}^{\mathrm{gt}} \mathbf{x}_{C,i} -
\mathbf{T}_{\mathrm{end}}^{\mathrm{gt}} \mathbf{x}_{C,i}
\right\|_2^2
}
\label{eq:ealign}
\end{equation}

PRISM-VO is first compared against the plenoptic \gls{vo} method SPO~\cite{zeller2018spo} using the dataset introduced in~\cite{zeller2018synchronized}. \cref{tab:evaluation_plenoptic} reports the number of sequences that satisfy high, medium, and coarse precision thresholds for the different metrics. This evaluation protocol is adopted because only per-sequence results are available for SPO.
Both methods recover the absolute scale accurately (scale close to 1), with PRISM-VO attaining slightly better results with $d_s' \leq 1.10$ on 8/11 sequences and $d_s' \leq 1.30$ for 10/11 sequences. More notably, PRISM-VO clearly reduces accumulated rotation error ($e_r < 4^\circ$ on 8/10 sequences) and lowers worst-case rotation. The scale drift is comparable to SPO with 7/11 sequences achieving $e_s' \leq 1.05$. The alignment error is also consistently smaller, especially at high and coarse precision. Additionally, PRISM-VO succeeds on one sequence where SPO fails, indicating greater robustness and stability beyond pure accuracy metrics. These results indicate that PRISM-VO provides more accurate and robust pose estimation, particularly in terms of alignment and rotational consistency, demonstrating the benefits of the proposed bundle-adjustment formulation for globally consistent \gls{vo}. Therefore, PRISM-VO establishes a new state-of-the-art in plenoptic camera-based \gls{vo}.

\begin{table}[tb]
\caption{\gls{vo} results for plenoptic methods on the dataset~\cite{zeller2018synchronized} over 11 sequences. The results are shown in terms of percentages of sequences reaching high/medium/coarse precision. The results for $d_s'$ and $e_s'$ are for a scale factor under 1.05/1.1/1.3, $e_\text{align}$ for values under 1~\%/2~\%/4~\%, $e_\text{r}$ for angles under 1\textdegree/2\textdegree/4\textdegree. Best results are in bold.}
\label{tab:evaluation_plenoptic}
\centering
\setlength{\tabcolsep}{10pt}
\begin{tabular}{ccccc}
\toprule
Method & \shortstack{Abs. scale\\error $d_s'$} & \shortstack{Scale\\drift $e_s'$} & \shortstack{Alignment\\error $e_\text{align}$} & \shortstack{Rotation\\error $e_\text{r}$} \\
\midrule
PRISM-VO (ours) 
& \textbf{5}/\textbf{8}/\textbf{10} 
& \textbf{7}/7/\textbf{10} 
& \textbf{4}/\textbf{6}/\textbf{10} 
& \textbf{6}/\textbf{8}/\textbf{10} \\

SPO~\cite{zeller2018spo}
& \textbf{5}/7/9 
& \textbf{7}/\textbf{8}/9 
& 2/\textbf{6}/9 
& 3/7/8 \\
\bottomrule
\end{tabular}
\end{table}

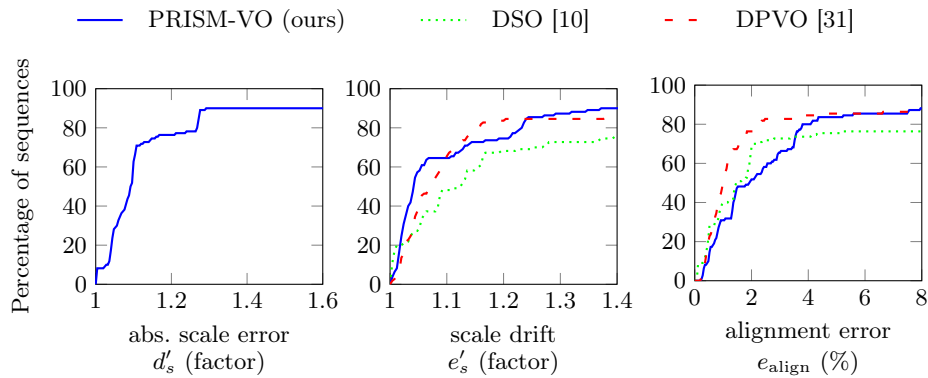
\begin{figure}[tb]
\centering

\pgfplotstableread[col sep=comma]{figures/cummulative_plots/files_csv/cum_abs_scale.csv}\absdata
\pgfplotstableread[col sep=comma]{figures/cummulative_plots/files_csv/cum_scale_drift.csv}\scaledriftdata
\pgfplotstableread[col sep=comma]{figures/cummulative_plots/files_csv/cum_alignment.csv}\alignmentdata

\colorlet{color_prism_vo}{blue}
\colorlet{color_dso}{green}
\colorlet{color_dpvo}{red}

\begin{tikzpicture}
\matrix[
    matrix of nodes,
    anchor=north west,
    inner sep=0.5em,
    column 2/.style={anchor=base west},
    column 5/.style={anchor=base west},
    column 8/.style={anchor=base west},
  ]
  {
      \ref{pgfplots:cumulative_prism-vo}& \small PRISM-VO (ours) &\quad\quad&
      \ref{pgfplots:cumulative_dso}& \small DSO~\cite{engel2018direct} &\quad\quad&
      \ref{pgfplots:cumulative_dpvo}& \small DPVO~\cite{teed2023deep} \\};
\end{tikzpicture}\\[-1ex]

\begin{subfigure}{0.34\textwidth}
  \centering
  \begin{tikzpicture}
\begin{axis}[
    width=3cm,
    scale only axis,
    xmin=1,
    xmax=1.6,
    xlabel={\shortstack{\small abs. scale error\\$d_s'$ (factor)}},
    ymin=0,
    ymax=100,
    ylabel={\small Percentage of sequences},
]

\addplot[color=color_prism_vo,solid,thick]
    table[x index=0,y index=1] {\absdata};
\end{axis}
\end{tikzpicture}
\end{subfigure}\hfill
\begin{subfigure}{0.31\textwidth}
  \centering
  \begin{tikzpicture}
\begin{axis}[
    width=3cm,
    scale only axis,
    xmin=1,
    xmax=1.4,
    xlabel={\shortstack{\small scale drift\\$e_s'$ (factor)}},
    ymin=0,
    ymax=100,
]
\addplot[color=color_prism_vo,solid,thick]
    table[x index=0,y index=1] {\scaledriftdata};
    \label{pgfplots:cumulative_prism-vo}
\addplot[color=color_dso,dotted,thick]
    table[x index=0,y index=2] {\scaledriftdata};
    \label{pgfplots:cumulative_dso}
\addplot[color=color_dpvo,loosely dashed,thick]
    table[x index=0,y index=3] {\scaledriftdata};
    \label{pgfplots:cumulative_dpvo}
\end{axis}
\end{tikzpicture}
\end{subfigure}\hfill
\begin{subfigure}{0.31\textwidth}
  \centering
  \begin{tikzpicture}
\begin{axis}[
    width=3cm,
    scale only axis,
    xmin=0,
    xmax=8,
    xlabel={\shortstack{\small alignment error\\$e_\text{align}$ (\%)}},
    ymin=0,
    ymax=100,
]
\addplot[color=color_prism_vo,solid,thick]
    table[x index=0,y index=1] {\alignmentdata};
\addplot[color=color_dso,dotted,thick]
    table[x index=0,y index=2] {\alignmentdata};
\addplot[color=color_dpvo,loosely dashed,thick]
    table[x index=0,y index=3] {\alignmentdata};
\end{axis}
\end{tikzpicture}
\end{subfigure}\hfill

\caption{Cumulative plots obtained on the dataset~\cite{zeller2018synchronized} for \gls{vo}/\gls{slam}.}
\label{fig:cumulative_error_plots}
\end{figure}

Apart from PRISM-VO and SPO, no other plenoptic \gls{vo} algorithms performing on this type of scene are available. Therefore, we also compare PRISM-VO with state-of-the-art monocular methods, namely DSO~\cite{engel2018direct}, ORB-SLAM3~\cite{campos2021orb} and DPVO~\cite{teed2023deep}. For a fair comparison of the tracking performance, large-scale-loop-closure is disabled in ORB-SLAM3. For the comparison methods, we use the provided monocular images and crop them to match the field of view of the plenoptic images (while adapting the intrinsics to preserve the pinhole model) to be able to compare the methods. To mitigate non-deterministic effects, each pipeline is run 10 times per sequence, with aggregated results shown in \cref{fig:cumulative_error_plots}. ORB-SLAM3 fails on all the sequences of \cite{zeller2018synchronized}, likely due to the limited field of view.
However, PRISM-VO is able to reliably recover the absolute scale of the scene.
Furthermore, PRISM-VO exhibits a consistently lower scale drift over the optimization-based method DSO and performs on par with the deep-learning-based approach DPVO.
With respect to the overall alignment error, PRISM-VO exhibits a performance similar to that of DSO and is only slightly worse than DPVO, which likely benefits from learned geometric priors.

\subsection{Quantitative Pose Estimation Error}

We further evaluate PRISM-VO on the LiFMCR dataset~\cite{FleithZirbel2025LiFMCR} to assess generalization to other plenoptic cameras. Raw images are calibrated with LiFCal~\cite{fleith2024lifcal}. The dataset only contains plenoptic data. For comparison with conventional methods, we generate totally focused images using a central perspective projection on a common image plane through the main lens center. This emulates a pinhole camera model and allows standard monocular methods to be applied (see supplementary material).

Ground-truth poses for all camera views are provided by a Vicon motion capture system, enabling per-frame pose evaluation. We evaluate absolute pose errors over the full trajectory and report the \gls{rmse} denoted as $\mathrm{RMSE}_t$ and $\mathrm{RMSE}_r$ for the translation and rotation errors respectively in \cref{tab:abs_error_lifmcr}. PRISM-VO achieves an error of a few millimeters and less than three degrees on most sequences. Although the camera model is not exactly a pinhole model for monocular methods, the comparison gives a rough idea. Overall, PRISM-VO consistently achieves the lowest translational error, obtaining the best $\mathrm{RMSE}_t$ on six out of seven sequences. Rotational accuracy is also competitive, achieving the best $\mathrm{RMSE}_r$ in four sequences and the second-best results in most remaining cases. The results demonstrate the advantages for plenoptic cameras in close range scenes, where reliable depth can be measured.

\begin{table}[tb]
\caption{Quantitative trajectory accuracy comparison on the LiFMCR dataset~\cite{FleithZirbel2025LiFMCR}. We report translational \gls{rmse} ($\mathrm{RMSE}_t$, [mm]) and rotational \gls{rmse} ($\mathrm{RMSE}_r$, [\textdegree]) for PRISM-VO and the reference methods. Best and second-best results per sequence are highlighted in bold and underlined, respectively.}
\label{tab:abs_error_lifmcr}
\centering
\begin{tabular}{llcccc}
\begin{tabular}{ccccccccc}
\toprule
 & \multicolumn{2}{c}{PRISM-VO (ours)} & \multicolumn{2}{c}{DSO} & \multicolumn{2}{c}{ORB-SLAM3 (mono)} & \multicolumn{2}{c}{DPVO} \\
Scene & $\mathrm{RMSE}_t$ & $\mathrm{RMSE}_r$ & $\mathrm{RMSE}_t$ & $\mathrm{RMSE}_r$ & $\mathrm{RMSE}_t$ & $\mathrm{RMSE}_r$ & $\mathrm{RMSE}_t$ & $\mathrm{RMSE}_r$ \\
\midrule
01 & \textbf{41.66} & \textbf{2.98} & \underline{59.32} & 48.56 & 271.24 & 98.79 & 607.44 & \underline{3.62} \\
02 & \textbf{57.81} & \underline{6.09} & \underline{112.56} & 21.24 & 316.78 & 25.66 & 356.60 & \textbf{2.60} \\
03 & \textbf{38.34} & 2.86 & 326.70 & \textbf{1.97} & \underline{123.20} & 10.71 & 169.23 & \underline{2.04} \\
04 & \textbf{10.53} & \textbf{2.83} & \underline{150.89} & 87.20 & 261.02 & 45.97 & 268.47 & \underline{3.01} \\
05 & \textbf{9.32} & \textbf{2.89} & 239.16 & 9.15 & \underline{121.64} & 56.93 & 219.66 & \underline{2.92} \\
06 & \textbf{12.21} & \textbf{3.31} & \underline{543.90} & 164.22 & 773.65 & \underline{153.28} & 704.67 & 163.09 \\
07 & \underline{110.87} & \underline{11.86} & 253.36 & 162.17 & 225.55 & 21.12 & \textbf{5.87} & \textbf{1.99} \\
\bottomrule
\end{tabular}
\end{tabular}
\end{table}

\subsection{Ablation Study}

We assess PRISM-VO components through incremental ablation, isolating each contribution and measuring its impact on accuracy and drift reduction.

\begin{itemize}
    \item \textbf{(1) Pinhole camera model:}  
    Baseline using a central perspective camera model without plenoptic geometry.

    \item \textbf{(2) Plenoptic depth initialization and model integration:} 
    Metric scale from the plenoptic depth and deep integration of the plenoptic camera model for the tracking and the bundle adjustment.

    \item \textbf{(3) Full PRISM-VO algorithm:}  
    Adds an inverse virtual depth residual term to the optimization. Automatically balances depth residuals by their uncertainty, assigning higher influence to points with lower variance.
\end{itemize}

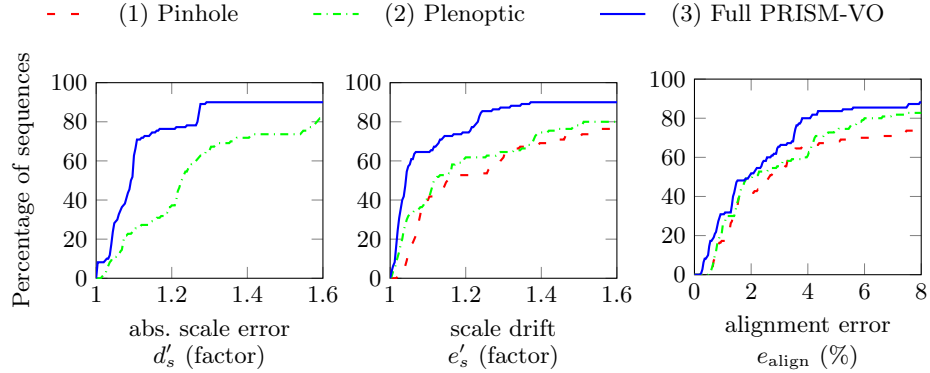
\begin{figure}[t]
\centering

\pgfplotstableread[col sep=comma]{figures/ablation_study/files_csv/cum_abs_scale.csv}\absdataablationstudy
\pgfplotstableread[col sep=comma]{figures/ablation_study/files_csv/cum_scale_drift.csv}\scaledriftdataablationstudy
\pgfplotstableread[col sep=comma]{figures/ablation_study/files_csv/cum_alignment.csv}\alignmentdataablationstudy

\colorlet{color_full_prism-vo}{blue}
\colorlet{color_plenoptic}{green}
\colorlet{color_pinhole}{red}

\begin{tikzpicture}
\matrix[
    matrix of nodes,
    anchor=north west,
    inner sep=0.5em,
    column 2/.style={anchor=base west},
    column 5/.style={anchor=base west},
    column 8/.style={anchor=base west},
  ]
  {
      \ref{pgfplots:AS_pinhole}& \small (1) Pinhole &\quad\quad&
      \ref{pgfplots:AS_plenoptic}& \small (2) Plenoptic &\quad\quad&
      \ref{pgfplots:AS_full_prism-vo}& \small (3) Full PRISM-VO \\};
\end{tikzpicture}\\[-1ex]

\begin{subfigure}{0.34\textwidth}
  \centering
  \begin{tikzpicture}

\begin{axis}[%
    width=3cm,
    scale only axis,
    xmin=1,
    xmax=1.6,
    xlabel={\shortstack{\small abs. scale error\\$d_s'$ (factor)}},
    ymin=0,
    ymax=100,
    ylabel={\small Percentage of sequences},
]

\addplot[color=color_plenoptic,dash dot,thick]
    table[x index=0,y index=2] {\absdataablationstudy};
\addplot[color=color_full_prism-vo, solid,thick]
    table[x index=0,y index=3] {\absdataablationstudy};
\end{axis}
\end{tikzpicture}
\end{subfigure}\hfill
\begin{subfigure}{0.31\textwidth}
  \centering
  \begin{tikzpicture}
\begin{axis}[%
    width=3cm,
    scale only axis,
    xmin=1,
    xmax=1.6,
    xlabel={\shortstack{\small scale drift\\$e_s'$ (factor)}},
    ymin=0,
    ymax=100,
]
\addplot[color=color_pinhole,loosely dashed,thick]
    table[x index=0,y index=1] {\scaledriftdataablationstudy};
    \label{pgfplots:AS_pinhole}
\addplot[color=color_plenoptic,dash dot,thick]
    table[x index=0,y index=2] {\scaledriftdataablationstudy};
     \label{pgfplots:AS_plenoptic}
\addplot[color=color_full_prism-vo, solid,thick]
    table[x index=0,y index=3] {\scaledriftdataablationstudy};
     \label{pgfplots:AS_full_prism-vo}
\end{axis}
\end{tikzpicture}
\end{subfigure}\hfill
\begin{subfigure}{0.31\textwidth}
  \centering
  \begin{tikzpicture}

\begin{axis}[%
    width=3cm,
    scale only axis,
    xmin=0,
    xmax=8,
    xlabel={\shortstack{\small alignment error\\$e_\text{align}$ (\%)}},
    ymin=0,
    ymax=100,
]
\addplot[color=color_pinhole,loosely dashed,thick]
    table[x index=0,y index=1] {\alignmentdataablationstudy};
\addplot[color=color_plenoptic,dash dot,thick]
    table[x index=0,y index=2] {\alignmentdataablationstudy};
\addplot[color=color_full_prism-vo, solid,thick]
    table[x index=0,y index=3] {\alignmentdataablationstudy};
\end{axis}
\end{tikzpicture}
\end{subfigure}\hfill

\caption{Cumulative error distributions for the ablation study. Curves compare the pinhole baseline (1), plenoptic model with depth initialization (2), and the full PRISM-VO system with variance-weighted depth and adaptive balancing (3). Results show progressive improvement, highlighting the benefit of tightly integrating plenoptic data.}
\label{fig:ablation_study}
\end{figure}

\cref{fig:ablation_study} reports the cumulative plots of the used metrics for the ablation stages.  Compared to the pinhole baseline (1), adding plenoptic depth initialization and explicit plenoptic projection (2) reduces scale drift and makes absolute scale observable, confirming the benefit of exploiting plenoptic data. The full PRISM-VO algorithm (3) further improves all metrics, most notably scale consistency and alignment, through variance-weighted inverse virtual depth residuals and adaptive residual balancing. This uncertainty-aware integration of plenoptic depth within bundle adjustment enhances global consistency and reduces accumulated drift. Overall, each component contributes positively, with the largest gains from jointly optimizing and probabilistically weighting depth information.

\section{Conclusion}
\label{sec:Conclusion}

We presented PRISM-VO, a novel \gls{vo} method that tightly integrates plenoptic camera measurements into a photometric bundle adjustment. By explicitly modeling the plenoptic geometry, the method jointly leverages photometric and depth information via an inverse virtual depth residual.
Experiments show improved trajectory consistency and accurate metric scale on challenging indoor and outdoor sequences. PRISM-VO outperforms the leading plenoptic VO pipeline and matches or surpasses monocular optimization and learning-based methods while reliably recovering metric scale. These results highlight the potential of tightly coupled plenoptic sensing and optimization for robust pose estimation and metric reconstruction with a single camera, despite limitations such as reduced resolution, a smaller field of view, and limited long-range depth accuracy.


\section*{Acknowledgements}
This research was partially funded by the Federal Ministry of
Research, Technology and Space of Germany in its program "FH-Kooperativ".

%
%
\bibliographystyle{splncs04}
\bibliography{main}
\end{document}